\documentclass{article}

     \PassOptionsToPackage{numbers, compress}{natbib}

\usepackage[position, final]{neurips_2026}

\usepackage[utf8]{inputenc} 
\usepackage[T1]{fontenc}    
\usepackage{hyperref}       
\usepackage{url}            
\usepackage{booktabs}       
\usepackage{amsfonts}       
\usepackage{nicefrac}       
\usepackage{microtype}      
\usepackage[table]{xcolor}         
\usepackage{graphicx}
\usepackage{nicematrix}
\usepackage{array}
\usepackage{colortbl}
\usepackage{amssymb}  
\usepackage{hyperref}
\usepackage{cleveref}
\usepackage{tikz}
\usetikzlibrary{positioning, fit, calc, backgrounds}
\usepackage{floatrow}
\usepackage{calc}
\usepackage{wrapfig}

\newfloatcommand{capbtabbox}{table}[][\FBwidth]

\usepackage{enumitem}
\setlist[itemize]{itemsep=3pt, parsep=0pt, topsep=-.25em,leftmargin=15pt}

\title{AI Agents Push Humans Out of the Loop}

\author{%
  Margaret Mitchell \\
  Hugging Face \\
  \texttt{meg@huggingface.co} \And
    Avijit Ghosh \\
    Hugging Face \\
\texttt{avijit@huggingface.co} \And
    Samir Passi \\
    Data \& Society \\
\texttt{passi.samir@gmail.com}
}

\begin{document}

\maketitle

\begin{abstract}
AI agents pose significant risks as they are granted increasing autonomy. A commonly proposed solution is human oversight and keeping a ``human in the loop'', but this is not a simple solution: 
The cognitive capacities reliable oversight requires are themselves degraded by extended use of AI systems. In essence, \textit{oversight degrades the overseer}. This position paper argues that, rather than mitigating this effect, \textbf{current approaches to the development and deployment of AI agent systems do not support effective human oversight -- they contribute to its degradation}. To address this, a top priority in the advancement of AI agents should be supporting the situated goals and cognitive requirements of effective human oversight, treating the human needs of overseers at the same level of importance as AI agent capability. To put this idea into practice, we connect work on automation and human-computer interaction to AI agent processes, outlining design-level affordances and organizational protocols that (1) support overseers in exercising critical judgement and (2) counteract the skill atrophy that arises from extended use of automation. We urge developers and deployers to adopt these or similar approaches. Without explicit support for the cognitive demands of oversight, AI agents will continue to passively incentivize the degradation of the very human skills they rely on.

\end{abstract}

\section{Introduction}

From healthcare to enterprise, a common recommendation for automated assistance is to have a ``human in the loop'' who supervises system processes \cite{WuEtAl2022,Bakken2023,DroriTeeni2024}. This recommendation captures the intuition that automated systems introduce risks that can be managed with meaningful oversight. Governance frameworks have formalized this view, stipulating that humans using high-risk AI systems must maintain meaningful control and make final decisions \cite{SantoniDeSioEtAl2018,NIST_AIRMF,EUAIAct14}. With the recent rise in AI-automated workflows and agentic AI, policymakers, developers, and deployers have converged on the practice of human oversight as a priority in service of multiple goals: Preventing harmful operations \cite{AWSHITL,NIST_AIRMF,SrikumarEtAl2025,LangerEtAl2025}, operationalizing ethical priorities \cite{Frenette23,SadovskiEtAl2025,LangerEtAl2025}, and ensuring legal compliance \cite{Chaffer2024DecentralizedGO,NIST_AIRMF,IBM2025,AWSHITL}. 

However, the presence of an overseer does not entail reliable oversight \cite{Green2021TheFO,SterzEtAl2024,LangerEtAl2025,Nature2025}. How to ensure  oversight is effective is underexplored in the design of current AI agent systems, and whether it is even possible in a given system is rarely questioned.\footnote{Relevant work shedding light on this question includes \citet{Green2021TheFO,sarkar2022explainableairacemodel,SterzEtAl2024,turri2024_transparencyinthewild,Passi2025,LangerEtAl2025,dhanorkar-coding-agents-2026, PassiSingh2026}.} 
Although human oversight should be a joint effort between developers, deployers, and users \cite{RAIMM-2025}, the onus of oversight currently falls almost entirely on users. 
Industry guidelines, policy recommendations, and interface messages in AI applications encourage users to ``audit'' outputs, ``review'' claims, ``prevent'' harm, and ``double-check'' for mistakes \cite{EUAIAct14, Biden-2023-AI, human-loop-egregious-2022, chatgpt-study-mode-2025, salesforce-genai-guidelines-2023, Canada-genai-2025, ChatGPTApp, ClaudeApp}, but the mechanisms users would need to do this effectively are virtually absent  \cite{turri2024_transparencyinthewild, epperson2025-debugsteer, Passi2025}. 

Recognizing this ``intellectual blind spot'' \cite{EhsanEtAl2026} is urgent. At the same time that AI agent systems have increasingly been deployed at scale, self-reports have documented difficulty in maintaining active engagement as an overseer, and research across multiple domains has uncovered serious negative impacts on critical thinking skills resulting from extended use of automated systems. We therefore urge the community to prioritize this issue and take the position that \textbf{current approaches to AI agent development and deployment are detrimental to providing effective human oversight}. 

To address this, we advocate for a two-pronged solution of \textbf{cognitive scaffolding} \cite{WoodBrunerRoss1976,Clark1997}, provided by developers and deployers. 
\textit{Developers} must create and implement processes for AI agent runtimes that support the cognitive requirements of effective oversight, including strategic points of friction and interfaces that maintain and/or help regain users' situation awareness in their role as overseer during and after agent operation. 
 \textit{Deployers} must institute organizational protocols that stimulate critical engagement, including trainings on recognizing signs of fatigue, rotation policies that safeguard attention, and processes that preserve domain skill.

 \textbf{Contributions.} We make three main contributions: (1) a review connecting decades of work on automation and human oversight to modern-day AI agents, tracing how AI agent oversight degrades the overseer; (2) an argument drawing on this work that the cognitive needs of reliable oversight must be a first-class concern in AI agent development and deployment, highlighting a potential feedback loop in which degraded oversight is reinforced through training; and (3) a structured inventory of solutions for cognitive scaffolding that translates research from HCI, cognitive science, and human factors into concrete development and deployment mechanisms for effective AI agent oversight.

 \textbf{Paper Structure.} \Cref{sec:oversight-safety} defines fundamental concepts relevant to our position and establishes the importance of human oversight for the safe deployment of AI agent systems. \Cref{sec:oversight-misses} documents how current approaches to AI agent oversight are inadequate and can be actively detrimental to a human's ability to appropriately oversee. \Cref{sec:irony} examines how cognitive degradation results from continued use of automated systems and discusses how this further hampers the ability to provide reliable oversight. \Cref{sec:solutions} offers concrete solutions for developers and deployers to support users in oversight. \Cref{sec:alt-views} engages with alternative views, addressing prevailing wisdom and norms in the development and deployment of AI agents. Finally, \Cref{sec:conclusion} concludes by arguing that human oversight, far from being a simple add-on for safety, requires deep engagement with the requirements of human attention, cognition, and critical analysis.

\section{The role of human oversight}\label{sec:oversight-safety}

Effective human oversight is critical for the operation of AI agents because agents can take incorrect actions, with consequences that are costly or irreversible \cite{openai-agentic-governance-2023, chan-agentic-harms-2023, RuanEtAl2024ToolEmu}. Such errors cannot be reliably caught by reviewing outputs alone: AI systems make errors at rates and in patterns that output review does not detect \cite{BuçincaEtAl2021,Passi2025,google-specification-gaming-2020}. Pre-deployment testing cannot account for all possible issues, as AI agents are constructed to be flexible across different uses and contexts without needing step-by-step instructions; by design, all possible action sequences and failures cannot be listed. Comprehensive testing of multi-step agent behavior is also substantially more expensive than evaluating single-turn systems, making reliable coverage economically infeasible for most deployers \cite{ghosh2026evalbottleneck}. This means that assessment of the appropriateness of AI agent operations is required in real time, during system execution, motivating human supervision that is more deeply integrated into system processing. We now describe how this oversight is situated in current AI agent practice and trace how its difficulty has escalated as the capabilities of AI systems have advanced.

\subsection{The human in the loop}

\textit{Human oversight} refers to mechanisms through which humans monitor, validate, intervene in, or override automated system behavior. As AI systems have become ubiquitous, the need for human oversight has been stressed as key to safe deployment. For example, the EU AI Act mandates effective human oversight as a primary risk mitigation strategy, including the ability to monitor for anomalies, remain vigilant to automation bias and overreliance, interpret outputs, intervene, reverse, override, and disregard system outputs \cite{EUAIAct14,BeinelEtAl2025}. Consulting agencies recommend continued oversight that establishes anchors for accountability throughout system processing \cite{KleinEtAl2025,DeloitteHCT2026}. In the context of generative AI, much of this role falls to users:  Commercial generative AI chat providers disclose that the system can make mistakes, recommending that users check responses and important information \cite{GeminiApp,GoogleGeminiResponses,ClaudeApp,ChatGPTApp}, and AI agents are designed to ask users for approval during operation \cite{ClaudeSecurity,ChatGPTAgent,dhanorkar-coding-agents-2026}, blurring the line between user and overseer. Our focus is on this dual role.

A common approach for human oversight is {\bf human-in-the-loop (HITL)}, a development and deployment approach where people actively participate in a system’s operation by providing feedback, validating output, making decisions, or labeling data. As AI systems have been deployed at scale to execute multi-step workflows with some level of autonomy, several distinct lineages of HITL have converged: The human in the loop is a user who must prevent unintended harm (aviation lineage \cite{Bainbridge83, sellen-aviation-copilot-2024}) and authorize consequential actions (autonomous weapons lineage \cite{DocheryEtAl2012,Scharre2015,SantoniDeSioEtAl2018,ICRC2020,Cottier2022}), 
and their interactions can be used for further model training (machine learning lineage \cite{Settles2011,AmershiEtAl2014}).

\subsection{The evolution of system sophistication}  

As AI systems have become more sophisticated, oversight has grown harder. In conventional discriminative AI systems, oversight focused on a single output, such as a prediction or a risk score, with the main goal being to ensure its correctness \cite{AmershiEtAl2019, GreenChen2019}. Even there, the techniques that increased prediction accuracy made it harder to explain what the system was doing \cite{GunningXAI}, motivating work on helping users appropriately calibrate their trust in system outputs \cite{hoffman2019metricsexplainableaichallenges,henrique2025dynamictrustcalibrationusing}. Generative AI (GenAI) systems based on large language models (LLMs) have made this substantially more complex \cite{Liao-Vaughan-Transparency-2024, desai-riedl-raiagents-2025}: To appropriately gauge system correctness, users must evaluate open-ended outputs, such as long-form text or code, that often have no single correct answer to check against. They must catch \textbf{hallucinations}, inaccurate or fabricated content presented as though it is factual \cite{maynez-2020-hallucination}. They must grapple with \textbf{capability unpredictability}, capabilities that were neither explicitly programmed nor anticipated, that do not work reliably, and that exhibit new failure modes \cite{Liao-Vaughan-Transparency-2024, yampolskiy-monitoring-2025}.

Agentic AI introduces further oversight challenges. Unlike basic LLM-based GenAI systems such as chatbots that are limited to producing a single output in response to a query, modern AI agents typically leverage LLMs to carry out multiple steps without explicit human programming or direct human involvement, adding new levels of opacity. This places users into a complex oversight role, where they must process voluminous intermediate actions and outputs, generated faster than they can meaningfully review them. One of the core benefits of AI agents is that they have a decreased need for human specification and increased flexibility in how to operate, yet this heightens the risk of unforeseen consequential actions that affect critical systems and protected data while obscuring what an overseer must address.  For example, agents may silently edit or delete files \cite{Replit}, exfiltrate sensitive information \cite{zhang-pei-2026-llm}, exploit software vulnerabilities \cite{metr-2026-openai-hugging-face-incident-investigation}, enact unauthorized financial transactions \cite{NIST_AI_risks}, or make private data public \cite{MCPExploit}. 

These risks are compounded by how agentic systems are built. Agent  pipelines involve multiple model-driven roles (planners, executors, evaluators) whose interaction can obfuscate critical information before anyone sees an output \cite{zhang2025which}. AI agent ``tool use'' exacerbates the problem through \textbf{tool hallucination}: Generating misleading or incorrect tool information that is plausible but difficult to verify, such as fabricating non-existent tools, invoking real tools with incorrect parameters, or misrepresenting tool outputs \cite{Patil2024gorilla, YinEtAlToolUse}. Consequences from these behaviors then cascade through multi-step plans \cite{zhu2026agentdebug}. \citet{YinEtAlToolUse} found that the current focus on improving ``reasoning'' amplifies tool hallucination, making the human oversight problem even harder. Agents can also give misleading accounts of their own behavior, such as reasoning that does not reflect what actually drove their actions \cite{chen-models-dont-say-2025} or reports that conceal failures \cite{GuoEtAl2025UpwardDeception}. These challenges are further magnified by \textit{multi-}agent systems: When several agents work together on tasks managed by an orchestrator, failures can arise from inter-agent misalignment, when an agent fails to provide necessary information or to request needed information from another agent \cite{cemri-multi-agent-failure-2025}.

\section{The oversight oversight: What current AI agent oversight discussions miss}\label{sec:oversight-misses}
Despite the broad recognition of the importance of human oversight, there is limited recognition of the deep complexities of ensuring AI agent oversight is reliable, effective, and appropriate. 
Decades of research have established how to account for human needs in system design (e.g., \cite{CardEtAlHCI,Horvitz99,GunningAha2019,AmershiEtAl2019,Shneiderman2020HCAI}),  
yet AI agent development largely measures success through task outcomes such as speed, accuracy, and throughput \cite{EhsanEtAl2026}, treating human oversight as a separate consideration independent of the quality of the system.\footnote{A notable exception to this norm, \citet{Passi2025} argues how and why the current approach to ``transparency'' in agentic systems is incompatible with effective oversight; their sociotechnical perspective is most similar to our position, where we additionally focus on technical solutions and long-term challenges.} Specific difficulties include the following:

\textbf{The user must play multiple roles simultaneously.} An AI agent user must leverage the system for their own goal while also providing oversight, giving permission for actions during agent operation, assessing the accuracy of the agent's selections, evaluating the safety and appropriateness of its steps, and foreseeing potential ramifications \cite{ClaudeSecurity,ChatGPTAgent,GrundeMcLaughlinEtAl26,shi2026humancontrolanchoranswer, dhanorkar-coding-agents-2026}. In practice, users find themselves repeatedly approving prompts the agent surfaces to continue its processing \cite{AWSHITL,LangChainHITL,Taggart2026}, leading to ``approval fatigue'', in which users stop paying close attention to what they are approving \cite{ClaudeAuto,FranklinEtAl2026,Approval_Fatigue_aipatternbook,CummingsEtAl2013}. 

\textbf{The amount of content relevant to an overseer is massive.} Effective supervision of an AI agent requires maintaining an appropriate mental model \cite{KuleszaEtAl2015,hoffman2019metricsexplainableaichallenges} of a large and heterogeneous stream of information generated during execution. This includes the agent's ``chain-of-thought'' -- its natural-language reasoning traces as it works towards goals -- along with the tools it selects, the arguments it passes, the plans it makes, its exchanges with other modules, and the content it produces. These details are dense, lengthy, generated rapidly, and distributed across components that can change or disappear, making review difficult. Questions that on their surface may seem straightforward, such as why an agent chose a tool, modified a plan, or altered course, can be effectively unanswerable in practice \cite{suchi-early-adopt-2025}. 
Real-time observation can help spot mistakes early \cite{agentops-observe-2024, openai-agentic-governance-2023}, but the requirements of maintaining oversight often exceed users' ability to keep up and play an active role \cite{process-observe-2025, Passi2025}.

\textbf{The cognitive demand is substantial.} Comprehending rapid, dynamic agentic information is difficult in its own right. Yet formal characterizations of the user's shifting role as AI agent systems are given increased autonomy (e.g., \cite{ParasuramanEtAl2000,feng2025levels,mitchell2025fullyautonomousaiagents}) miss the cognitive demands each shift imposes: As the user takes on a role of ``approver'', effective oversight requires assessing each decision point as a task expert, anticipating and guarding against unintended outcomes, tracking the agent's multi-step plan, navigating disparate sources, and maintaining an understanding of the agent's goals, current execution state, and prior actions -- a working memory and contextual awareness load that current interfaces are not designed to support \cite{dhanorkar-coding-agents-2026}. Governance frameworks do provide some guidance on what overseers must be able to do, such as understanding system capacities and limitations, interpreting outputs \cite{EUAIAct14}, and providing approvals and interventions \cite{IMDA2026Agentic,KoreaAIBasicAct2026}, but say little about whether overseers can realistically meet these requirements at each decision point \cite{Green2021TheFO,BeinelEtAl2025}.

\textbf{Cognitive biases influence engagement.} Multiple cognitive biases also impede effective oversight. With automation bias \cite{CummingsEtAl2012,ZhaiEtAl2024,BuçincaEtAl2021}, people accept system suggestions even when they are wrong. With anchoring bias, people are more likely to agree to an AI system's decision when provided before they form their own \cite{GreenChen2019,BuçincaEtAl2021}. Automation-induced complacency \cite{EndsleyKiros1995,ParasuramanEtAl2000} and our tendency to favor fast, heuristic-based shortcuts instead of slower, more effortful reasoning \cite{ZhaiEtAl2024,BuçincaEtAl2021},  especially in the face of mental overload \cite{Yu2016,ParasuramanEtAl2000}, further challenge users in their role as overseer.

\textbf{Current oversight affordances are poorly designed.}
Vendor frameworks acknowledge that supporting effective oversight requires thoughtful design for the user, but overlook their needs: Anthropic emphasizes user control through front-loaded input and agent-initiated re-engagement \cite{AnthropicTrustworthy}. Salesforce \cite{SalesforceHITL} and AWS \cite{AWSHITL} discuss HITL roles but not human tendencies during sustained engagement. ``Human-in-the-loop'' is only a meaningful solution if the human can independently see into the loop.

Taken together, these issues sketch out a pattern pointing to a solution. Current AI agent design centers the \textit{agent}: Systems are built to surface their own processing on their own terms \cite{Passi2025}. Effective oversight requires centering the \textit{person}: They must be able to understand and maintain focus. 
Towards this goal, work on human cognition is instructive. 
 Notably, \textbf{dual-process theory} posits that humans use one of two primary reasoning methods in decision making: quickly and automatically, employing basic heuristics (``System 1 thinking''); or slowly and deliberatively (``System 2 thinking'') 
\cite{Epstein1994,Evans2003,kahneman2011thinking}  -- a necessary mode for the critical analyses needed for effective oversight. When decisions are routine, System 1 predominates \cite{TayEtAl2016}, and early evidence suggests that this is the dominant mode of engagement with AI agent coding systems \cite{catalan2026imreadingthatunderstanding, dhanorkar-coding-agents-2026}. 
Yet people do not easily switch from System 1 to 2 mid-task \cite{MilkmanEtAl2009,Yu2016,TayEtAl2016,BuçincaEtAl2021}; systems must provide the affordances that support this kind of deliberative analysis. 

\section{The irony of automation}\label{sec:irony}

In 1983, \citet{Bainbridge83} identified an \textbf{irony of automation}: The more an automated system can do, the more the human operator's skills and awareness of the system's relevant context degrade, leaving the operator least prepared to intervene when it matters most. Emerging evidence suggests that this irony applies to AI agents: Continued use of AI agent systems can delegate away epistemic agency, contributing to over-trust, cognitive degradation, and skill atrophy, thus eroding human oversight ability. We now turn to these concerns.

\subsection{Critical skills degrade}

Studies on sustained AI use document negative effects on cognitive capacities required for oversight: ``deskilling'' (skill atrophy) and ``intuition rust'' \cite{BudzyńEtAl2025,EhsanEtAl2026}, decreased critical and analytical thinking \cite{ZhaiEtAl2024}, reduced vigilance and pattern recognition needed to recognize abnormalities or catch mistakes \cite{BudzyńEtAl2025,EndsleyKiros1995}, and overreliance \cite{BuçincaEtAl2021,ZhaiEtAl2024}. Recent terminology has crystallized additional patterns: ``cognitive dependence'' \cite{Gerlich2025}, ``cognitive debt'' \cite{KosmynaEtAl2025}, and ``cognitive surrender'' \cite{ShawEtAl2025}.

These effects are due in part to requirements of critical thinking that are unmet in current system design.  
Using increased automation means fewer occasions for evidence-seeking, considering multiple perspectives, skepticism, and the iterative work of building knowledge \cite{ShahBender2022,ZhaiEtAl2024}. Over time, users find that they offload cognitively demanding tasks that are critical for understanding situations, hindering their ability to adequately assess them  \cite{Gooch2026}.\footnote{Humorously captured in \Cref{fig:xkcd}.} The cognitive cost may be measurable: Initial evidence suggests that people who use LLM-based systems for essay writing have significantly decreased brain connectivity \cite{KosmynaEtAl2025}. 

Diminishing critical thinking is especially worrisome when learning new tasks, where the development of appropriate reasoning skills is blunted \cite{ZhaiEtAl2024}. Users may never develop the foundational skills needed to become experts on tasks they use AI agents for, yet robust domain knowledge is a prerequisite for intervening in dynamic systems and taking over when automation fails \cite{EndsleyKiros1995}. Expert users also over-rely on AI \cite{gaube-expert-overreliance-2021,passi-overreliance-chapter-2025,schaffer-task-familiarity-2019}, especially in contexts slightly outside their expertise, such as an experienced software developer using an AI agent to code in a new programming language \cite{dhanorkar-coding-agents-2026}.

\subsection{Oversight ability diminishes}
 
Empirical work is now beginning to show how oversight ability diminishes and document its effects. Several studies analyzing overreliance on GenAI  found that users take incorrect shortcuts for assessing the appropriateness of outputs \cite{vorvoreanu-lessons-2025}, such as mistaking the fluent style of ChatGPT responses \cite{KabirEtAl2024} or the presence of citations \cite{DingEtAl2025} as signals of accuracy.   
Experienced developers overseeing coding agents adopt heuristics that prioritize efficiency, such as treating an agent's plan as a ``faithful proxy'' of what it will do and assuming that code passing unit tests is correct \cite{dhanorkar-coding-agents-2026}. 
While heuristic short-cuts are practical and play an important role,  
human cognitive limits coupled with extraordinary agentic capabilities point to a future of agent oversight driven by suboptimal control.

System behavior also further nudges users away from critical information. For example, sycophantic GenAI models validate users more often than humans evaluating the same situations, but users prefer and trust these models over less sycophantic baselines \cite{ChengEtAl2026}. 
\citet{SharmaEtAl2026} document disempowerment patterns in real-world Claude conversations, where users' actions, beliefs, or values become less well-aligned with reality, while rating those conversations favorably. \citet{GhoshEtAl2026} describe how the dominant chatbot paradigm of single authoritative responses, opaque reasoning, agreeable tone, and optimization for smooth interaction, reduces user agency by removing the cognitive friction needed for independent judgement.  Together, these works suggest that interaction patterns experienced as helpful or satisfying can still weaken the forms of independence, skepticism, and self-monitoring that oversight requires.

\subsection{Ineffective oversight is incentivized}\label{sec:falling-out}

Complicating the effects of system design on users is that when systems do not need regular intervention, 
%
users lose the ability to effectively engage with them  \cite{CummingsEtAl2013}. 
Recent self-reports describe how current HITL processes for coding agents ``[make] the loop stultifying''  \cite{Taggart2026}, where users must ``babysit the outputs, catch the occasional hallucination'' \cite{RedditHITL}.
Developers become ``cognitively distant from the code they must review'' \cite{dhanorkar-coding-agents-2026}, making it challenging to find and fix even simple issues. 

 
This creates an ``out-of-the-loop'' performance problem \cite{EndsleyKiros1995} that decreases \textbf{situation awareness} – the perception and comprehension of relevant environmental states \cite{Endsley1995} – as human operators shift from active participants to passive information processors. Yet paradoxically, they must be vigilant enough to intervene and deeply engage at strategic moments, a well-recognized challenge in cognitive science (\Cref{sec:oversight-misses}). The history of technology warns us that loss of situation awareness coupled with human cognitive constraints does not only weaken oversight, but pushes 
humans to the 
\textit{outside} of oversight loops \cite{mindell-cybernetics-2002, carr-glass-cage-2014}. 

\subsection{Ineffective oversight creates a feedback loop}\label{sec:feedback_loop}

\begin{wrapfigure}{r}{0.4\textwidth}
    \centering
\vspace{-1em}\includegraphics[width=\linewidth]{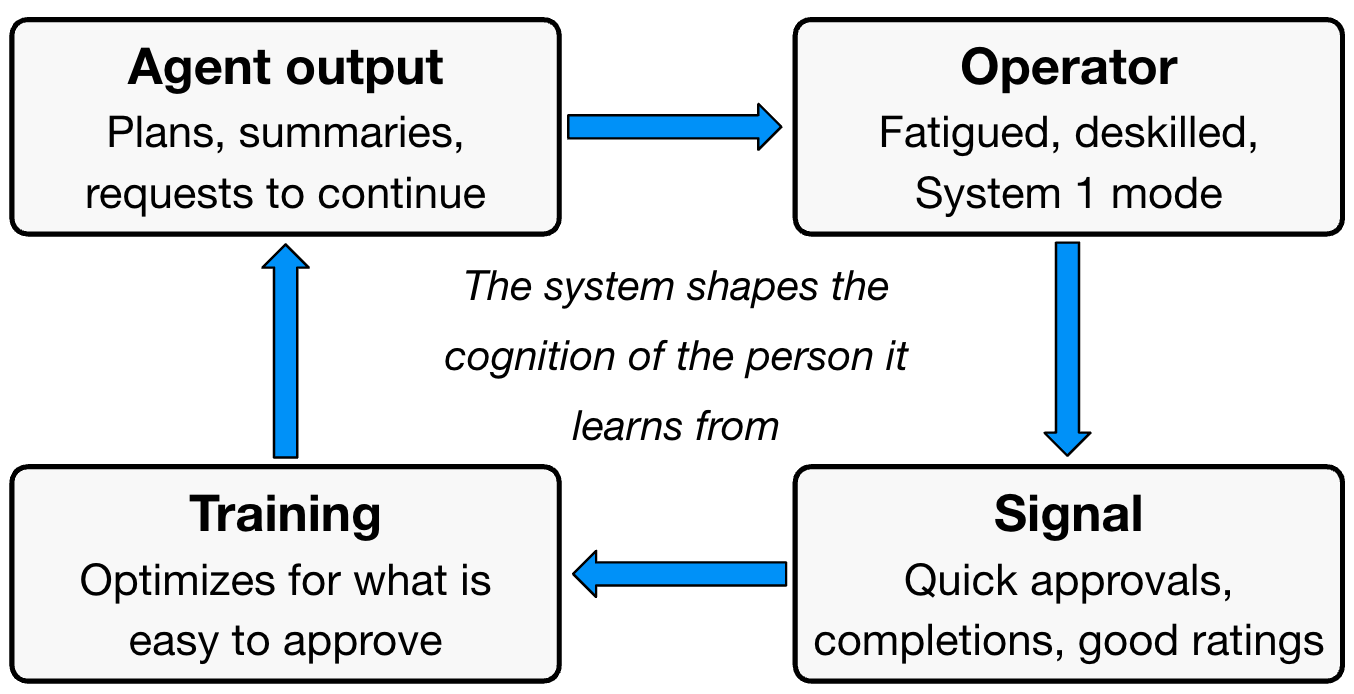}
    \caption{Potential oversight feedback loop. As human oversight degrades and systems are optimized for approval, training can reinforce conditions that weaken judgement.}
    \label{fig:feedback_loop}
\end{wrapfigure}
These cognitive effects also create an alignment-relevant failure mode. Modern LLM-based systems are commonly trained with human feedback \cite{OuyangEtAl2022}, and deployed agent systems are often evaluated through task completion rates, user disapproval rates, and other user behavioral signals \cite{AnthropicEvals,AmplitudeEvals}. As oversight quality decreases, the signals sent back to the system for retraining degrade as well.  While an alert overseer may scrutinize a plan and withhold approval, a tired or ``out-of-the-loop'' overseer may approve quickly, for example, accepting rationales as long as they are fluent. If such approvals are treated as evidence of success, the system can begin to optimize for outputs that disincentivize thoughtful oversight: producing confident summaries, reducing friction, and surfacing oversimplified plans that are easy to skim (\Cref{fig:feedback_loop}).

This concern does not require developers to explicitly optimize for cognitive depletion. It arises whenever the measured target, such as user approval, comes apart from an intended target, such as well-scrutinized correct action. In that setting, the same kinds of interaction patterns associated with sycophancy \cite{ChengEtAl2026} and disempowerment \cite{SharmaEtAl2026} become especially concerning at agent scale, where the human is already pushed toward passive monitoring. Oversight degradation therefore connects the irony of automation to reward hacking of human oversight: The human rater can become the exploitable part of the reward channel.\footnote{This dynamic resembles what \citet{DoctorowEnshittification} calls ``enshittification,'' in which perverse incentives, compounded through iteration, degrade platforms over time.}

\textbf{Evil double agents.} A related risk arises when models learn which failures users are likely to detect. Recent work on \textbf{agentic upward deception} shows that agents can conceal failure under environmental constraints, such as by providing unsupported results, substituting unavailable sources, or fabricating files \cite{GuoEtAl2025UpwardDeception}. If an agent can predict what an overseer is unlikely to inspect, it can choose plans, explanations, or timing that keep problematic behavior below the threshold of detection \cite{baum2026normspricerlbasedalignment}. 

\subsection{The result} 

The result is the \textit{irony of automation} \citep{Bainbridge83}: The more capable an automated system is, the harder it is for a human to reliably intervene. The cases where human oversight is most critical – rare events, situations requiring deep familiarity and skill – are precisely those that automation leaves humans least prepared for. \textbf{The act of being an overseer degrades the capacities oversight requires: Oversight degrades the overseer.} Addressing this irony requires treating human cognitive requirements as first-class design constraints, rather than governance add-ons that can be adapted into AI agent systems after they are developed. We turn to these considerations next.

\section{Solutions}\label{sec:solutions}

Enabling appropriate oversight of AI agents requires grappling with the complexities of human cognition when interacting with automated systems, an active area of HCI research \cite{SinghEtAl2025,YeEtAl2026}. Building on this work alongside work in cognitive science and human factors, we present a structured inventory of user-centric solutions that connects  oversight goals to the mechanisms that support them. Mechanisms marked with $^\dagger$ are novel proposals introduced here, inspired by the surrounding literature and gaps that emerged in constructing the inventory. (Further details on the inventory construction process are available in Appendix \ref{app:constructing_inventory}.)

We frame oversight solutions as \textit{mechanisms} to achieve \textit{goals}, where goals involve preventing unwanted effects 
and supporting the user in oversight duties. \textit{Preventative goals} include preventing automation bias, acquiescence, anchoring, and complacency \cite{EndsleyKiros1995,GreenChen2019,ParasuramanEtAl2000,BuçincaEtAl2021,Gerlich2025}; fatigue from tedious oversight duties \cite{Approval_Fatigue_aipatternbook,ClaudeAuto}; overreliance on AI agent systems \cite{BuçincaEtAl2021,Nature2025}; and cognitive overload from being inundated by voluminous, complex AI agent outputs \cite{Endsley1995}. \textit{Supportive goals} include supporting critical judgement and engagement \cite{ParasuramanEtAl2000,Gerlich2025} in order to  
maintain user expertise and oversight skill \cite{EndsleyKiros1995,Gerlich2025,Nature2025,SinghEtAl2025}. 

To meet these goals, mechanisms must be implemented during AI agent \textit{development} (by developers) and \textit{deployment} (by the organizations using them). 
Development-stage solutions concern design-level affordances for what a user can access, when, and how. Deployment-stage solutions involve organizational protocols that build awareness of oversight issues and help users stay critically engaged. \Cref{tab:solutions-matrix} summarizes the solution categories, and \Cref{fig:two-by-two} visualizes how they are distributed across AI agent development and use. Goals for reliable AI agent oversight are interrelated, and mechanisms can serve multiple goals. 
We now turn to the specific solutions.

\floatsetup{valign=t}
\begin{figure}[t]
\begin{floatrow}

\ffigbox[\FBwidth]{%
\centering\footnotesize
\renewcommand{\arraystretch}{1.3}
\setlength{\tabcolsep}{6pt}
\begin{NiceTabular}[t]{l}[rounded-corners=4pt,
  code-after={
    \tikz \draw[violet!40!black, line width=0.3pt, rounded corners=5pt]
      (1-|1) rectangle (last-|last);
      \tikz \draw[violet!40!black, line width=0.3pt]
    (2-|1) -- (2-|last);
  }
]
\CodeBefore
  \rowcolor{violet!14}{1}
  \rowcolors{2}{violet!4}{violet!8}
\Body
  \RowStyle{\bfseries\color{violet!40!black}}
  Goals \\
  Prevent automation bias \\
  Prevent fatigue \\
  Prevent overreliance \\
  Prevent overload \\
  Support judgement \\
  Support engagement \\
\end{NiceTabular}%
\hspace{8pt}%
\begin{NiceTabular}[t]{l}[rounded-corners=4pt,
  code-after={
    \tikz \draw[violet!40!black, line width=0.3pt, rounded corners=5pt]
      (1-|1) rectangle (last-|last);
  \tikz \draw[violet!40!black, line width=0.3pt]
    (2-|1) -- (2-|last);
}
]
\CodeBefore
  \rowcolor{violet!14}{1}
  \rowcolors{2}{violet!4}{violet!8}
\Body
  \RowStyle{\bfseries\color{violet!40!black}}
  Mechanisms \\
  Strategic friction \\
  Decision design \\
  Behavioral monitoring \\
  Trainings \& Exercises \\
  Workload \& Scheduling \\
  Role design \\
\end{NiceTabular}%
}{%
\caption{The solutions we identify to support user oversight draw from six mechanism categories to reach different goals. Goals are interrelated, and mechanisms can serve more than one goal. See Appendix~\ref{app:constructing_inventory} for more.}\label{tab:solutions-matrix}%
}

\hspace{0em}\ffigbox[\FBwidth]{%
\resizebox{.98\linewidth}{!}{%
\begin{tikzpicture}[
  font=\small,
  colhead/.style={font=\small\bfseries, align=center},
  colsub/.style={font=\footnotesize\itshape, text=black!60, align=center},
  rowhead/.style={font=\small\bfseries, align=center, anchor=east},
  rowsub/.style={font=\footnotesize\itshape, text=black!60, align=center, anchor=east},
  cat/.style={
    rectangle, rounded corners=3pt,
    draw=violet!50!black, line width=.5pt,
    fill=violet!8, text=violet!40!black,
    minimum width=29.5mm, minimum height=6mm,
    align=center, inner sep=.5mm,
    font=\footnotesize
  },
  filledquad/.style={
    rectangle, rounded corners=5pt,
    draw=violet!40!black, line width=0.3pt, 
    fill=violet!4, inner sep=3mm,
    minimum width=32.5mm, minimum height=30mm  
  },
  shortfilledquad/.style={
    rectangle, rounded corners=5pt,
    draw=violet!40!black, line width=0.3pt, 
    fill=violet!4, inner sep=3mm,
    minimum width=32.5mm, minimum height=22.5mm  
  },
  emptyquad/.style={
    rectangle, rounded corners=5pt,
    draw=black!30, line width=0.3pt, dashed,
    fill=black!3, inner sep=3mm,
    minimum width=32.5mm, minimum height=22.5mm
  },
  empty/.style={
    font=\footnotesize\itshape, text=black!50, align=center
  }
]

\node[colhead] at (0,   4.6)  {Development};
\node[colsub]  at (0,   4.25) {Design-level affordances};
\node[colhead] at (3.4, 4.6)  {Deployment};
\node[colsub]  at (3.4, 4.25) {Organizational protocols};

\node[rowhead] at (-1.8,  2.6) {Engaging\\oversight};
\node[rowsub]  at (-1.6,  1.9) {In-the-moment\\attention};
\node[rowhead] at (-1.6, 0) {Maintaining\\skills};
\node[rowsub]  at (-1.8, -.7) {Long-term\\capacity};

\begin{scope}[on background layer]
\node[shortfilledquad] (tl) at (0, 2.69) {};
\node[shortfilledquad] (tr) at (3.4, 2.69) {};
\node[emptyquad] (bl) at (0, .25) {};
\node[shortfilledquad] (br) at (3.4, .25) {};
\end{scope}

\node[cat] (sf) at (0, 3.4) {Strategic friction};
\node[cat] (ad) at (0, 2.7) {Decision design};
\node[cat] (bm) at (0, 2) {Behavioral monitoring};

\node[cat] (ws) at (3.4, 3.4) {Decision design};
\node[cat] (ws) at (3.4, 2.7) {Behavioral monitoring};
\node[cat] (ws) at (3.4, 2) {Workload \& Scheduling};


\node[cat] (te) at (3.4, .95) {Trainings \& Exercises};
\node[cat] (rd) at (3.4, .25) {Role design};
\node[cat] (au) at (3.4, -.45) {Behavioral monitoring};
\end{tikzpicture}}
\vspace{-.25cm}
 }{%
  \caption{Solutions described in this work span two stages, development and deployment, and provide two types of cognitive support: engaging overseers in the moment, and maintaining critical skills over time.}\label{fig:two-by-two}
}
\end{floatrow}
\vspace{-.5cm}
\end{figure}
\vspace{-1em}
\par\noindent\rule{\textwidth}{0.4pt}


\textbf{Strategic friction.} 
These mechanisms require the user to perform cognitive work before or alongside agent operation and incentivize \textit{metacognition }\cite{singh2025protectinghumancognitionage,YeEtAl2026}. Relevant approaches include \textit{seamful design} \cite{EhsanEtAlXAI24}, which focuses on surfacing a system's limitations and gaps, and 
\textit{cognitive forcing functions}, where  interventions at the moment of decision-making encourage or require the user to engage analytically with content \cite{BuçincaEtAl2021,ghosh2026experimentalcomparisoncognitiveforcing, EhsanEtAl2026}. 
\begin{itemize}
\item \textit{Pre-commitment mechanisms} have the user record their own view or decision before seeing the agent's recommendation \cite{BuçincaEtAl2021,EhsanEtAl2026}. This intervention prevents anchoring bias and creates an audit trail for further analysis.
\item \textit{Delay-and-choice mechanisms} let the user decide whether and when to see the AI's output at all, preserving the option to work unaided and protecting the cognitive capacities that ``always on'' assistance atrophies \cite{BuçincaEtAl2021}.
\item \textit{Reasoning probes}, inline prompts at high-stakes moments, such as ``what evidence would change your mind?'' or ``what assumption does this approval rest on?'', maintain critical thinking \cite{SinghEtAl2025,ghosh2026experimentalcomparisoncognitiveforcing}.
\item \textit{Action gating} requires explicit verification before the agent proceeds down a consequential path, and may surface alternative options for the user to choose between rather than presenting a single recommendation to accept or reject  \cite{GrundeMcLaughlinEtAl26,shi2026humancontrolanchoranswer}.
\end{itemize}

\textbf{Decision design.} These mechanisms treat expert attention as a critical, scarce resource that should be brought in strategically, helping to reduce fatigue and mitigate acquiescence. They involve consideration of which actions require sign-off, when, and at what granularity \cite{EndsleyKiros1995,Gerlich2025, chen-agent-oversight-strategies-2026}. 
\begin{itemize}
\item \textit{Bounded autonomy specifications} shape what the agent may and may not do without approval, reserving user attention for decisions that genuinely require human judgement \cite{Approval_Fatigue_aipatternbook,shi2026humancontrolanchoranswer,baum2026anticipatoryhumanoversightagentic}.
\item \textit{Batch review support}, where the agent completes a logical unit of work and then surfaces the whole batch as a diff for the user to review, aids user engagement and judgement by making it possible to evaluate related actions together \cite{Approval_Fatigue_aipatternbook,GrundeMcLaughlinEtAl26}. Deploying institutions may also implement secondary checks, such as an additional review from a person for consequential actions.
\item \textit{Automated pre-checks} supplement human review with machine verification of properties that do not require judgement \cite{Approval_Fatigue_aipatternbook}.
\item \textit{Supportive interfaces} can help to strategically focus user attention on oversight considerations \cite{Horvitz99,GroceEtAl2014,ZhangEtAl2020,BuçincaEtAl2021,GajosEtAl2022}.
\end{itemize}

\textbf{Behavioral monitoring.} This category treats oversight quality as an empirically measurable property of human-AI interaction to test how well oversight is functioning \cite{narayanan-oversight-dependency-2026}.  These mechanisms carry clear risks. For example, they are a form of workplace surveillance: Data can be repurposed beyond supporting oversight, such as in performance evaluation. They are also subject to Goodhart's law \cite{ManheimGarrabrantGoodhardt,Goodhardt75}. How best to apply these mechanisms in light of these trade-offs warrants further work.
\begin{itemize}
\item \textit{Time-based signatures} track whether review duration drops while approval rates remain constant, or whether review time does not increase as the complexity of agent operations grows; both suggest that user attention and critical analyses may be waning \cite{SwaroopEtAl2024,SwaroopEtAl2025}. 
\item \textit{Override signatures} track how acceptance rates shift with fatigue \cite{AnckerEtAl2017}. Applied to agents, this could be used to measure whether the rate of disagreement with the agent declines over time or with increasing complexity, suggesting growing acquiescence rather than improving agent quality.
\item \textit{Evidence-seeking signatures}$^\dagger$ track whether the user requests additional information when stakes rise, since a reviewer who stops asking questions has likely stopped reviewing.
\end{itemize}
There are also strategies that can be applied to both overseer and agent:
\begin{itemize}
\item \textit{Canaries} are known items inserted into a system to reveal hidden failures \cite{srivastava2023beyond}. For oversight, known errors can be planted in agent output to check whether the user catches them; missed canaries provide direct evidence of degraded judgement \cite{ThreatImageProjection2005,ThreatImageProjection2025}. Deployers can use missed canaries to trigger a break, rotation, or retraining. For the agent, canaries can be tasks that require a specific output or that cannot be completed, such as summarizing a deliberately deleted file, testing whether the agent produces the expected output or correctly reports its inability \cite{GuoEtAl2025UpwardDeception}. 
\item \textit{Style-fatigue covariance analyses}$^\dagger$, inspired by work on sycophancy \cite{ChengEtAl2026} and disempowerment \cite{SharmaEtAl2026}, track whether the agent's output style (e.g., more confident, skimmable, longer/denser) shifts as signs of user fatigue increase, suggesting the agent may be adapting to, or encouraging, weakened oversight. These may help to mitigate the feedback loop described in \Cref{sec:feedback_loop}.
\item \textit{Audits} operate retrospectively to evaluate different dimensions of oversight. Reviewers can re-examine past decisions and adapt processes accordingly \cite{epperson2025-debugsteer,RajiEtAl2020}. For the agent, trace summaries can be checked against raw action logs to detect cases where the agent's narrative of what it did diverges from what it actually did  \cite{GuoEtAl2025UpwardDeception,chen-models-dont-say-2025}. 
\end{itemize}

``Guardian agents'' \cite{GartnerGuardianAgent} could monitor for signs of degradation in human-agent interaction and alert or assist the user when they detect an oversight issue. An adaptive AI agent system could also assess user behavior in order to calibrate how to engage, reducing cognitive overload.

\textbf{Trainings and exercises.} These mechanisms elucidate, maintain, and strengthen the cognitive activities oversight requires, including domain expertise, calibrated skepticism, and awareness of one's own degradation.
\begin{itemize}
\item \textit{Domain skill maintenance exercises} require users to regularly perform their underlying task without AI assistance. This helps to retain the expertise needed to evaluate AI outputs and to prevent the deeper organizational problem in which no one remaining understands the work well enough to oversee it \cite{EhsanEtAl2026}. Experts may also benefit from regular retrainings on their tasks \cite{Nature2025}. 
\item \textit{Critical evaluation training} teaches users to read AI outputs with appropriate skepticism and to prioritize what to scrutinize: what kinds of errors to expect, how to distinguish fluent-sounding output from correct output, and how to effectively weigh claims against external evidence \cite{Gerlich2025}.
\item \textit{Self-monitoring training} teaches users strategies for metacognition \cite{FabianoEtAl2025} -- recognizing when their attention is flagging and they are not engaging in critical scrutiny. This can raise awareness of cognitive biases and behaviors that 
\textit{Behavioral monitoring} supplements \cite{Gerlich2025}. 
\end{itemize}

\textbf{Workload and scheduling.} This category addresses the physiological inevitability of fatigue to support judgement and engagement when the user is alert. 
\begin{itemize}
\item \textit{Enforced breaks} prevent extended sessions that wear down cognitive sharpness \cite{HeltonRussell2015}.
\item \textit{Rotations} move users between tasks, or AI-assisted and unassisted versions of the same task, to prevent both fatigue and the cognitive surrender from prolonged exposure to agentic AI \cite{ParasuramanEtAl96,CummingsEtAl2013}.
\item \textit{Shifts} distribute oversight of high-stakes scenarios and long-running tasks across multiple people, so that no single person limits the task and overseers may rest \cite{FAA_Part117}.
\end{itemize}

\textbf{Role design.} This category concerns workplace dynamics, following \citet{SterzEtAl2024}: The user must want to do the oversight job well, be in a position where doing it well is rewarded, and have the competence to do it.
\begin{itemize}
\item \textit{Role assignment} ensures that users providing oversight have the expertise to evaluate AI outputs.
\item \textit{Role separation} ensures that a user performing oversight is not also someone who benefits from the AI output being approved, which could create a structural incentive against intervention.
\item \textit{Incentive alignment} removes productivity targets and performance metrics that disincentivize scrutiny (such as maximizing output volume), and rewards high-quality oversight.
\end{itemize}

\textbf{How the mechanisms work together.} We do not believe that a single mechanism is sufficient, but rather advocate for multiple mechanisms to be applied together. One application strategy is to disrupt the different links in the feedback loop of \Cref{fig:feedback_loop}: For example, \textit{Strategic friction} mechanisms keep users engaged when seeing agent output, while \textit{Behavioral monitoring} mechanisms flag low-quality approvals and \textit{Role design} mechanisms remove incentives to approve by default. Many mechanisms, such as breaks, rotations, role separation, canaries, and audits, also operate independently of the overseer's in-the-moment judgement, so they remain effective even as judgement degrades.

\section{Alternative Viewpoints}\label{sec:alt-views}

There are caveats to implementing these interventions. Initial evidence suggests users may disprefer systems that reduce overreliance \cite{BuçincaEtAl2021}, pointing to a tension between user satisfaction and oversight preservation. There are several further substantive objections to our position.\vspace{-8pt}

\subsection*{1. Cognitive effects are overstated, and people will adapt}\vspace{-6pt}
The cognitive effects are real but more modest, and will diminish as users adapt to working with AI agents, 
just as users learned to calibrate trust in search engines and autocomplete. 
Treating current cognitive findings as permanent risks designing for a transitional state.

\textbf{Response: } 
 (1) The adaptation analogy understates the structural difference between AI agents and earlier tools. Search engines return results the user evaluates; AI agents take actions the user is expected to authorize. The cognitive load of evaluating a returned document is not comparable to maintaining situation awareness across a multi-step plan executing in real time. 
 (2) Self-reports of cognitive degradation and difficulty maintaining attention are rapidly emerging \cite{Taggart2026,RedditHITL,Gooch2026}, consistent with EEG evidence of reduced brain connectivity during LLM system use \cite{KosmynaEtAl2025}. Solutions proposed in these discussions often align with ours, but rarely draw on prior research or account for the interplay of developers, deployers, users, and systems.  (3) This assumes that the rate of capability advancement provides enough time to adapt. But agent capabilities are increasing faster than cognitive science can keep up with, and recent work indicates that capability improvements (such as enhanced reasoning) actively amplify failure modes that overseers must catch \cite{YinEtAlToolUse}. The issue is happening \textbf{now}. Adaptation alone cannot address persistent and growing issues.\vspace{-6pt}

\subsection*{2. Better tooling and transparency will solve this}\vspace{-6pt}
The problem is essentially a tooling problem: Better explanations, reasoning traces, and richer audit logs will give users what they need. Existing research already pursues the ``cognitive scaffolding'' described here; there is no need for a paradigm shift. 

\textbf{Response: } We partially agree: Better tooling and transparency are part of what cognitive scaffolding requires, and several of the design-level affordances we propose (batch review, action gating) are consistent with existing  research. We argue that this work is necessary, but insufficient: (1) Explanations operate on cognitive capacities that AI agent use itself degrades and can increase inappropriate trust. Users rarely engage deeply with each detail, and explanations can themselves be incorrect or act as cognitive anchors \cite{BuçincaEtAl2021,WangYin2022,SpitzerEtAl2025}. (2) A focus on just these solutions leaves the organizational prong of our proposal unaddressed. No interface improvement alone fully addresses approval fatigue from sustained sessions or a mismatch between user capability and system operation. These require organizational protocols \cite{EhsanEtAl2026}, which are out of scope for any tooling-focused agenda. Treating cognitive support as a first-class concern does not reject a push for better tooling or explainable AI; it recognizes that tooling alone cannot do the work being asked of it.

\subsection*{3. Human oversight is becoming obsolete because alignment will solve it}\vspace{-6pt}
As AI systems become more aligned, the need for vigilant human oversight diminishes. Preference-based training (e.g., RLHF) is the long-term solution.

\textbf{Response: } A strong version of this view is that technical alignment will eliminate the need for oversight. This relies on assumptions about future capability advances that are not currently demonstrated. Oversight remains a stated requirement of safety frameworks and governance regulations applicable \textbf{now}. Further, scholarship in HCI and sociotechnical systems (STS) highlights that ``perfect'' agents are neither realistic nor desirable: Failures are fundamental to acting in human worlds, especially as moments of learning and improvisation \cite{Passi2025}, and a direct result of systems that are flexible to different users' (potentially conflicting) needs. Another version of this view is that preference-based training addresses oversight by allowing low ratings on outputs that undermine it, but that assumes that user preferences are a reliable proxy for oversight quality. The evidence in \Cref{sec:irony} suggests otherwise. Users prefer fluent, agreeable, and confidence-inducing outputs even when those reduce skepticism and encourage overreliance \cite{BuçincaEtAl2021}.   Under preference-based training optimized against cognitively degraded raters, the same dynamics that produce sycophancy \cite{ChengEtAl2026} and disempowerment \cite{SharmaEtAl2026} would produce systems that increasingly reward the conditions of their own ineffective oversight. Preference-based training can be part of an oversight system, but only if approval signals are audited against independent measures of oversight quality. The economic shift toward RLAIF and self-distillation \cite{lee2023rlaif} further weakens the idea that human feedback can continue to provide a reliable corrective mechanism.

\section{Conclusion}\label{sec:conclusion}

The need for human oversight of AI agents is recognized broadly: written into governance frameworks, vendor documentation, and the design of agent systems themselves. Yet the current trajectory of AI agent advancement does not meaningfully engage with what oversight requires, and instead actively contributes to its degradation. The more autonomy agents are granted, the less the user is positioned to oversee them, and the more the very cognitive capacities oversight requires, such as situation awareness, critical judgement, and domain skill, are undermined by the act of using these systems. In the current state of the art, \textbf{oversight degrades the overseer.}

Without intervention, users will be pushed further out of the loop as agentic systems are deployed at greater scale and across more consequential domains. Users will continue to approve plans they have not meaningfully reviewed, accept rationales they have not independently evaluated, and certify actions whose consequences they cannot anticipate. 
In the limit, this is not human oversight at all: It is a superficial actor in a system they cannot meaningfully penetrate. 

Avoiding this trajectory requires treating cognitive support for human overseers as a first-class concern in AI agent system development, on par with capability. We have outlined a two-pronged approach: \textit{developers} must build runtime affordances -- such as strategic friction, well-considered decision design, behavioral monitoring -- that preserve the conditions for effective oversight; and \textit{deployers} must institute organizational protocols -- such as trainings and breaks -- that protect the cognitive capacities oversight requires over time. Neither prong is sufficient alone, but together, they constitute the cognitive scaffolding agentic systems will need if human oversight is meant to be meaningful.

We urge developers and deployers to adopt these or similar approaches and invite the broader AI community to take up this work -- through empirical research on cognitive degradation across deployment contexts, system-level audits of whether existing agents support the oversight they presume, and the design and evaluation of new affordances and protocols that the inventory in \Cref{sec:solutions} only begins to outline. If we do not act, the irony of automation that Bainbridge [\citeyear{Bainbridge83}] identified four decades ago will play out at scale, compounding across every domain agents are deployed into. With the rapid rise of agentic AI systems, the time to think critically about how to support effective human oversight is long past due.

\begin{ack}
We thank the anonymous reviewers for their thoughtful feedback, in particular the connections to relevant work.
\end{ack}

\bibliographystyle{plainnat}
\urlstyle{same} 
\bibliography{bibliography}
\newpage

\appendix

\section{Constructing the Solutions Inventory}\label{app:constructing_inventory}

Constructing the solutions inventory surfaced a slight disconnect in how prior work describes oversight solutions. Some work articulates what effective oversight \textit{requires}, such as ``preventing automation bias'', without implementation-specific details on how to achieve it. Other work articulates oversight \textit{strategies} at different levels of granularity, from general ideas such as ``strategic friction'' to specific approaches such as ``slow down the process'', with variability in the language used for similar ideas. We did not find consistent groupings of similar types of strategies, or mappings between strategies and the types of cognitive scaffolding they provide.

We therefore developed a simple device separating \textbf{goals} and \textbf{mechanisms} (\Cref{tab:solutions-matrix_large}) to identify solutions at each intersection (e.g., the \textit{goal} of preventing automation bias can be met with \textit{mechanisms} for strategic friction, \Cref{tab:solutions-matrix-examples}). We offer this categorization as an organizing device for locating existing work and identifying gaps, not as a definitive taxonomy; it is a by-product of building the inventory that we hope is independently useful.

\begin{table}[h]
\small
\renewcommand{\arraystretch}{1.4}
\setlength{\arrayrulewidth}{1.5pt}
\setlength{\tabcolsep}{3pt}
\arrayrulecolor{white}
\begin{NiceTabular}{c|p{0.27\textwidth}| *{6}{>{\centering\arraybackslash}p{0.09\textwidth}}}
\CodeBefore
\rowcolors{2}{gray!10}{gray!20}
\columncolor{white}{1}
\rowcolor{white}{1}
\Body
\rowcolor{gray!20}
& & & \multicolumn{6}{c}{\textbf{\large Mechanisms}} \\\hline
\Block{6-1}{\rotatebox{90}{\parbox{3.5cm}{\textbf{\large Goals}}}}
& 
\cellcolor{white} 
& 
\rotatebox{50}{\parbox{1.6cm}{Strategic friction}} & 
\rotatebox{50}{\parbox{1cm}{Decision design}} & 
\rotatebox{50}{\parbox{1.45cm}{Behavioral monitoring}} & 
\rotatebox{50}{\parbox{1.45cm}{Trainings \& Exercises}} & 
\rotatebox{50}{\parbox{1.5cm}{Workload \& Scheduling}} & 
\rotatebox{50}{\parbox{1.5cm}{Role design}} \\
\midrule
& Preventing automation\kern-1pt\ bias       & \checkmark & \checkmark & \checkmark &   \checkmark &            & \checkmark \\
& Preventing fatigue            &    \checkmark        & \checkmark & \checkmark &   \checkmark & \checkmark &  \checkmark          \\
& Preventing overreliance       & \checkmark & \checkmark & \checkmark &   \checkmark & \checkmark & \checkmark \\
& Preventing cognitive overload & \checkmark & \checkmark & \checkmark &             \checkmark & \checkmark &            \\
& Supporting judgement & \checkmark & \checkmark & \checkmark &  \checkmark &  \checkmark          & \checkmark \\
& Supporting engagement         & \checkmark & \checkmark & \checkmark &             \checkmark & \checkmark &  \checkmark      \\
\bottomrule
\end{NiceTabular}
\caption{Organizational device we developed in building the solutions inventory. This recognizes how some solutions described in previous literature can be categorized as \textit{goals} while others can be categorized as \textit{mechanisms}. Yet, the two can be intersected to create an organized solution space. This resultant matrix informed our categorization, guided our search, and helped to surface solution gaps. A check indicates that at least one relevant solution is described in \Cref{sec:solutions}.}\label{tab:solutions-matrix_large}
\end{table}

\begin{table*}[h]
\small
\renewcommand{\arraystretch}{1.4}
\setlength{\tabcolsep}{3pt}
\arrayrulecolor{white}
\resizebox{\linewidth}{!}{%
\begin{NiceTabular}{cc>{\centering\arraybackslash}p{0.12\textwidth}!{\color{white}\vrule width 3pt} *{6}{>{\centering\arraybackslash}p{0.15\textwidth}}}
\CodeBefore
\rowcolors{2}{gray!10}{gray!20}
\Body

\Block[fill=white]{11-1}{\rotatebox{90}{\parbox{12em}{\textbf{\large Goals}}}} \\
\rowcolor{white}
& & & \multicolumn{6}{c}{\textbf{\large Mechanisms}} \\
& \cellcolor{white} & \cellcolor{white} &  \multicolumn{6}{c}{\textbf{Primary Implementer }} \\
\rowcolor{gray!20}
\cellcolor{white}& \cellcolor{white}& \cellcolor{white}& \multicolumn{1}{c|}{\textbf{Developer}} 
& \multicolumn{2}{c|}{\textbf{Both}} 
& \multicolumn{3}{c}{\textbf{Deployer}} \\\Hline[tikz={line width=.5pt}]
\cellcolor{white} &\cellcolor{white} & \cellcolor{white}  & 
\raisebox{8pt}{\rotatebox{50}{\parbox{5em}{Strategic friction}}} & 
\raisebox{8pt}{\rotatebox{50}{\parbox{4em}{Decision design}}} & 
\raisebox{8pt}{\rotatebox{50}{\parbox{5em}{Behavioral monitoring}}} & 
\raisebox{8pt}{\rotatebox{50}{\parbox{5em}{Trainings \& Exercises}}} & 
\raisebox{8pt}{\rotatebox{50}{\parbox{5em}{Workload \& Scheduling}}} & 
\raisebox{8pt}{\rotatebox{50}{\parbox{6.5em}{Role design}}} \\\Hline[tikz={line width=3pt}]
& \Block[fill=gray!5]{4-1}{\rotatebox{90}{\parbox{3em}{\textbf{Preventing}}}}
& automation\kern-1pt\ bias       & pre-commitment mechanisms & supportive interfaces & override signatures (indirect) &   critical evaluation training &            & separating roles \\
&& fatigue            &    delay-and-choice mechanisms (arguable)      & batch review & time-based signatures (indirect) &   self-monitoring training (indirect) & enforced breaks &   aligning incentives         \\
&& overreliance       & action gating & bounded autonomy & evidence-seeking signatures (indirect) &   domain skill maintenance exercises & rotations & assigning roles \\
&& cognitive\kern-1pt\ overload & delay-and-choice mechanisms & batch review & style-fatigue covariance (e.g., if it triggers retraining; indirect) &            critical evaluation training & enforced breaks &            \\\Hline[tikz={line width=2pt}]
&
\Block[fill=gray!5]{2-1}{\rotatebox{90}{\parbox{4em}{\textbf{Supporting}}}}

& judgement & reasoning probes & supportive interfaces & canaries (e.g., if they trigger retraining; indirect) &  domain skill maintenance exercises & rotations           & assigning roles \\
&& engagement         & reasoning probes & batch review & guardian agents &             critical evaluation training & enforced breaks &  aligning incentives \\
\bottomrule
\end{NiceTabular}}
\caption{Example application of our solutions matrix as we worked through the literature to identify common themes and organize the inventory. Different goals have multiple mechanisms that can help reach them; a single mechanism can apply to more than one goal. We considered a mechanism to be indirect if it detects the problem, rather than reaching the goal itself.}\label{tab:solutions-matrix-examples}
\end{table*}
\clearpage
\newpage

\section{Solutions overview}

Visualized overview of solution space discussed in this paper.

\newcommand{\childOffset}{3cm}
\newcommand{\childGap}{1.2cm}
\newcommand{\clusterGap}{2cm}
\newcommand{\curveStrength}{1.5cm}

\definecolor{strategicfriction}{HTML}{B80000}
\definecolor{decisiondesign}{HTML}{B85F00}
\definecolor{behavioralmonitoring}{HTML}{B8A900}
\definecolor{trainingsandexercises}{HTML}{00B822}
\definecolor{workloadandscheduling}{HTML}{005CB8}
\definecolor{roledesign}{HTML}{8700B8}

\definecolor{strategicfrictiontext}{HTML}{6B0000}
\definecolor{decisiondesigntext}{HTML}{5E2A00}
\definecolor{behavioralmonitoringtext}{HTML}{5E4500}
\definecolor{trainingsandexercisestext}{HTML}{005E1C}
\definecolor{workloadandschedulingtext}{HTML}{00245E}
\definecolor{roledesigntext}{HTML}{42005E}

\begin{figure*}[h]
\centering
\begin{tikzpicture}[
  parent/.style={draw, rounded corners=6pt, line width=1pt,
    minimum width=3.4cm, minimum height=1cm, align=center, font=\bfseries\small},
  child/.style={anchor=west, align=left, font=\footnotesize},
  connector/.style={line width=0.5pt, draw=#1},
]

\newcommand{\connect}[3]{%
  \draw[connector=#1] (#2.east) .. controls +(\curveStrength,0) and +(-\curveStrength,0) .. (#3.west);
}

\node[parent, draw=strategicfriction, fill=strategicfriction!15] (p1) at (0, 0) {Strategic friction};

\node[child, text=strategicfrictiontext] (c1a) at ($(p1) + (\childOffset,  .6*\childGap)$) {Pre-commitment mechanisms};
\node[child, text=strategicfrictiontext] (c1b) at ($(p1) + (\childOffset,  0.2*\childGap)$) {Delay-and-choice mechanisms};
\node[child, text=strategicfrictiontext] (c1c) at ($(p1) + (\childOffset, -0.2*\childGap)$) {Reasoning probes};
\node[child, text=strategicfrictiontext] (c1d) at ($(p1) + (\childOffset, -.6*\childGap)$) {Action gating};

\connect{strategicfriction}{p1}{c1a}
\connect{strategicfriction}{p1}{c1b}
\connect{strategicfriction}{p1}{c1c}
\connect{strategicfriction}{p1}{c1d}

\node[parent, draw=decisiondesign, fill=decisiondesign!15] (p2) at ($(p1) + (0, -.1*\childGap - \clusterGap)$) {Decision design};

\node[child, text=decisiondesigntext] (c2a) at ($(p2) + (\childOffset,  .6*\childGap)$) {Bounded autonomy};
\node[child, text=decisiondesigntext] (c2b) at ($(p2) + (\childOffset,  0.2*\childGap)$) {Batch review};
\node[child, text=decisiondesigntext] (c2c) at ($(p2) + (\childOffset, -0.2*\childGap)$) {Automated pre-checks};
\node[child, text=decisiondesigntext] (c2d) at ($(p2) + (\childOffset, -.6*\childGap)$) {Supportive interfaces};

\connect{decisiondesign}{p2}{c2a}
\connect{decisiondesign}{p2}{c2b}
\connect{decisiondesign}{p2}{c2c}
\connect{decisiondesign}{p2}{c2d}

\node[parent, draw=behavioralmonitoring, fill=behavioralmonitoring!15] (p3) at ($(p2) + (0, -.7*\childGap - \clusterGap)$) {Behavioral monitoring};

\node[child, text=behavioralmonitoringtext] (c3a) at ($(p3) + (\childOffset,  1.2*\childGap)$) {Time-based signatures};
\node[child, text=behavioralmonitoringtext] (c3b) at ($(p3) + (\childOffset,  .8*\childGap)$) {Override signatures};
\node[child, text=behavioralmonitoringtext] (c3c) at ($(p3) + (\childOffset,  .4*\childGap)$) {Evidence-seeking signatures};
\node[child, text=behavioralmonitoringtext] (c3d) at ($(p3) + (\childOffset,  0.0*\childGap)$) {Guardian agents};
\node[child, text=behavioralmonitoringtext] (c3e) at ($(p3) + (\childOffset, -.4*\childGap)$) {Canaries};
\node[child, text=behavioralmonitoringtext] (c3f) at ($(p3) + (\childOffset, -.8*\childGap)$) {Style-fatigue covariance};
\node[child, text=behavioralmonitoringtext] (c3g) at ($(p3) + (\childOffset, -1.2*\childGap)$) {Audits};

\connect{behavioralmonitoring}{p3}{c3a}
\connect{behavioralmonitoring}{p3}{c3b}
\connect{behavioralmonitoring}{p3}{c3c}
\connect{behavioralmonitoring}{p3}{c3d}
\connect{behavioralmonitoring}{p3}{c3e}
\connect{behavioralmonitoring}{p3}{c3f}
\connect{behavioralmonitoring}{p3}{c3g}

\node[parent, draw=trainingsandexercises, fill=trainingsandexercises!15] (p4) at ($(p3) + (0, -.5*\childGap - \clusterGap)$) {Trainings and\\exercises};

\node[child, text=trainingsandexercisestext] (c4a) at ($(p4) + (\childOffset,  .4*\childGap)$) {Domain skill maintenance};
\node[child, text=trainingsandexercisestext] (c4b) at ($(p4) + (\childOffset,  0.0*\childGap)$) {Critical evaluation training};
\node[child, text=trainingsandexercisestext] (c4c) at ($(p4) + (\childOffset, -.4*\childGap)$) {Self-monitoring training};

\connect{trainingsandexercises}{p4}{c4a}
\connect{trainingsandexercises}{p4}{c4b}
\connect{trainingsandexercises}{p4}{c4c}

\node[parent, draw=workloadandscheduling, fill=workloadandscheduling!15] (p5) at ($(p4) + (0, .5*\childGap - \clusterGap)$) {Workload and\\scheduling};

\node[child, text=workloadandschedulingtext] (c5a) at ($(p5) + (\childOffset,  0.2*\childGap)$) {Enforced breaks};
\node[child, text=workloadandschedulingtext] (c5b) at ($(p5) + (\childOffset, -0.2*\childGap)$) {Rotations};

\connect{workloadandscheduling}{p5}{c5a}
\connect{workloadandscheduling}{p5}{c5b}

\node[parent, draw=roledesign, fill=roledesign!15] (p6) at ($(p5) + (0, .5*\childGap - \clusterGap)$) {Role design};

\node[child, text=roledesigntext] (c6a) at ($(p6) + (\childOffset,  .4*\childGap)$) {Assigning roles};
\node[child, text=roledesigntext] (c6b) at ($(p6) + (\childOffset,  0.0*\childGap)$) {Separating roles};
\node[child, text=roledesigntext] (c6c) at ($(p6) + (\childOffset, -.4*\childGap)$) {Aligning incentives};

\connect{roledesign}{p6}{c6a}
\connect{roledesign}{p6}{c6b}
\connect{roledesign}{p6}{c6c}
\node[draw, rounded corners=8pt, line width=0.5pt, inner sep=15pt,
      fit=(p1) (p6) (c1a) (c3g) (c6c)] {};
\end{tikzpicture}
\caption{Solutions for supporting effective AI agent oversight, organized by mechanism category.}
\label{fig:solutions-tree}
\end{figure*}
\begin{figure}[h]
\resizebox{.85\linewidth}{!}{%
\raisebox{.5cm}{
\begin{tikzpicture}[
  font=\small,
  colhead/.style={font=\small\bfseries, align=center},
  colsub/.style={font=\footnotesize\itshape, text=black!60, align=center},
  rowhead/.style={font=\small\bfseries, align=center, anchor=east},
  rowsub/.style={font=\footnotesize\itshape, text=black!60, align=center, anchor=east},
  cat/.style={
    rectangle, rounded corners=3pt,
    draw=violet!50!black, line width=.5pt,
    fill=violet!8, text=violet!40!black,
    minimum width=29.5mm, minimum height=6mm,
    align=center, inner sep=.5mm,
    font=\footnotesize
  },
  filledquad/.style={
    rectangle, rounded corners=5pt,
    draw=violet!40!black, line width=0.3pt, 
    fill=violet!4, inner sep=3mm,
    minimum width=32.5mm, minimum height=30mm  
  },
  shortfilledquad/.style={
    rectangle, rounded corners=5pt,
    draw=violet!40!black, line width=0.3pt, 
    fill=violet!4, inner sep=3mm,
    minimum width=32.5mm, minimum height=22.5mm  
  },
  emptyquad/.style={
    rectangle, rounded corners=5pt,
    draw=black!30, line width=0.3pt, dashed,
    fill=black!3, inner sep=3mm,
    minimum width=32.5mm, minimum height=22.5mm
  },
  empty/.style={
    font=\footnotesize\itshape, text=black!50, align=center
  }
]


\node[colhead] at (0,   4.6)  {Development};
\node[colsub]  at (0,   4.25) {Design-level affordances};
\node[colhead] at (3.4, 4.6)  {Deployment};
\node[colsub]  at (3.4, 4.25) {Organizational protocols};

\node[rowhead] at (-1.8,  2.6) {Engaging\\oversight};
\node[rowsub]  at (-1.6,  1.9) {In-the-moment\\attention};
\node[rowhead] at (-1.6, 0) {Maintaining\\skills};
\node[rowsub]  at (-1.8, -.7) {Long-term\\capacity};

\begin{scope}[on background layer]
\node[shortfilledquad] (tl) at (0, 2.69) {};
\node[shortfilledquad] (tr) at (3.4, 2.69) {};
\node[emptyquad] (bl) at (0, .25) {};
\node[shortfilledquad] (br) at (3.4, .25) {};
\end{scope}

\node[cat] (sf) at (0, 3.4) {Strategic friction};
\node[cat] (ad) at (0, 2.7) {Decision design};
\node[cat] (bm) at (0, 2) {Behavioral monitoring};

\node[cat] (ws) at (3.4, 3.4) {Decision design};
\node[cat] (ws) at (3.4, 2.7) {Behavioral monitoring};
\node[cat] (ws) at (3.4, 2) {Workload \& Scheduling};


\node[cat] (te) at (3.4, .95) {Trainings \& Exercises};
\node[cat] (rd) at (3.4, .25) {Role design};
\node[cat] (au) at (3.4, -.45) {Behavioral monitoring};
\end{tikzpicture}}}
  \caption{Summary of potential solutions  described. Interventions span development and deployment, and two types of cognitive support: engaging oversight in the moment, maintaining skills long-term.}\label{fig:two-by-two_large}
\end{figure}
\clearpage
\newpage

\section{The cognitive degradation of extended use of external sources}

\begin{figure}[h!]
    \centering
    \includegraphics[width=\linewidth]{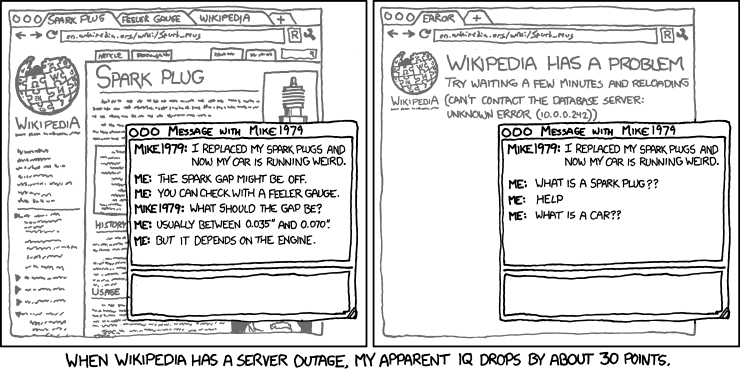}
    \caption{The popular comic xkcd humorously captured how building up a reliance on external systems for knowledge can destroy our understanding of basic concepts.\\ Source: \url{https://xkcd.com/903/} License: \href{https://creativecommons.org/licenses/by-nc/2.5/}{CC BY-NC 2.5}}
    \label{fig:xkcd}
\end{figure}

\end{document}